\documentclass[10pt,twocolumn,letterpaper]{article}

\usepackage{cvpr}
\usepackage{times}
\usepackage{epsfig}
\usepackage{graphicx}
\usepackage{amsmath}
\usepackage{amssymb}
\usepackage{color}
\usepackage{multirow}
\usepackage{caption}
\usepackage{subcaption}
\usepackage{wrapfig}
\usepackage[ruled,lined]{algorithm2e}
\usepackage{algorithmic}
\usepackage{lipsum}
\usepackage{enumitem}
\usepackage{paralist}
\usepackage{cuted}

\newcommand\blfootnote[1]{%
  \begingroup
  \renewcommand\thefootnote{}\footnote{#1}%
  \addtocounter{footnote}{-1}%
  \endgroup
}

\usepackage[pagebackref=true,breaklinks=true,letterpaper=true,colorlinks,bookmarks=false]{hyperref}

\cvprfinalcopy %

\def\cvprPaperID{8448} %

\newcommand{\name}{StarNet}
\newcommand{\datasetname}{Waymo Open Dataset}

\ifcvprfinal\fi
\begin{document}

\title{StarNet: Targeted Computation for\\Object Detection in Point Clouds}

\author{
  Jiquan Ngiam$^{*\dagger}$, Benjamin Caine$^{*\dagger}$, Wei Han$^{\dagger}$, Brandon Yang$^{\dagger}$,\\
  Yuning Chai$^{\ddagger}$, Pei Sun$^{\ddagger}$, Yin Zhou$^{\ddagger}$, Xi Yi, Ouais Alsharif$^{\ddagger}$, Patrick Nguyen,\\
  Zhifeng Chen$^{\dagger}$, Jonathon Shlens$^{\dagger}$, Vijay Vasudevan$^{\dagger}$\\
  $^{\dagger}$Google Brain, $^{\ddagger}$Waymo\\
  \texttt{\{jngiam,bencaine\}@google.com}\\
}

\maketitle

\begin{abstract}
Detecting objects from LiDAR point clouds is an important component of self-driving car technology as LiDAR provides high resolution spatial information. Previous work on point-cloud 3D object detection has re-purposed convolutional approaches from traditional camera imagery. 
In this work, we present an object detection system called \name\ designed specifically to take advantage of the sparse and 3D nature of point cloud data. \name\ is entirely point-based, uses no global information, has data dependent anchors, and uses sampling instead of learned region proposals. We demonstrate how this design leads to competitive or superior performance on the large \datasetname\ \cite{waymo_open_dataset} and the KITTI \cite{geiger2013vision} detection dataset, as compared to convolutional baselines. In particular, we show how our detector can outperform a competitive baseline on Pedestrian detection on the \datasetname\ by more than 7 absolute mAP while being more computationally efficient. We show how our redesign---namely using only local information and using sampling instead of learned proposals---leads to a significantly more flexible and adaptable system: we demonstrate how we can vary the computational cost of a single trained \name\ without retraining, and how we can target proposals towards areas of interest with priors and heuristics. Finally, we show how our design allows for incorporating temporal context by using detections from previous frames to target computation of the detector, which leads to further improvements in performance without additional computational cost.
\end{abstract}
\vspace{-10pt}
\section{Introduction}
Detecting and localizing objects forms a critical component of any autonomous driving platform \cite{geiger2013vision, nuscenes2019}.
As a result, self-driving cars (SDC) are equipped with a variety of sensors such as cameras, LiDARs, and radars \cite{cho2014multi, thrun2006stanley}, which the perception system must use to create an accurate 3D representation of the world.
Due to the nature of the driving task, the perception system must operate in real-time and in a highly variable operating environment \cite{kim2013parallel}.
LiDAR is one of the most critical sensors as it natively provides high resolution, accurate 3D data about the environment.
However, LiDAR based object detection systems for SDCs look remarkably similar to systems designed for generic camera imagery.
\ifcvprfinal\blfootnote{$^{*}$ Denotes equal contribution and authors for correspondence. JN proposed the idea and implemented the model. BC, JN, BY, YC, ZF and VV developed the infrastructure and experimented with the model. WH, VV, PS, JS built the evaluation framework.  XY, YZ, PN and OA developed early pieces of infrastructure and the dataset. JS, JN, VV, BC and others wrote the manuscript.}\fi

Object detection research has matured for camera images with systems evolving to solve camera-specific challenges such as multiple overlapping objects, large intra-class scale variance due to camera perspective, and object occlusion \cite{girshick2014rich, girshick2015fast, faster_rcnn, lin2016feature, lin2017focal}.
These modality-specific challenges make the task of localizing and classifying objects in imagery uniquely difficult, 
as an object may occupy any pixel, and neighboring objects may be as close as one pixel apart.
This necessitates treating every location and scale in the image equally, which naturally aligns with the use of convolutional networks for feature extraction \cite{girshick2014rich,girshick2015fast}. 
While convolutional operations have been heavily optimized for parallelized hardware architectures, scaling these methods to high resolution images is difficult as computational cost scales quadratically with image resolution.

In contrast, LiDAR is naturally sparse; 3D objects have real world scale with no perspective distortions, and rarely overlap.
Additionally, in SDC perception, every location in the scene is not equally important \cite{zeng2019end, bojarski2016end, bansal2018chauffeurnet}, and that importance can change dynamically based on the local environment and context.
Despite large modality and task-specific differences, the best performing methods for 3D object detection re-purpose camera-based detection architectures.
Several methods apply convolutions to discretized representations of point clouds in the form of a projected Birds Eye View (BEV) image \cite{yang2018pixor,luo2018fast,yang2018hdnet,lang2018pointpillars}, or a 3D voxel grid  \cite{zhou2018voxelnet, yan2018second}.
Alternatively, methods that operate directly on point clouds have re-purposed two stage object detector design, replacing feature extraction operations but still adopting the same camera-inspired region proposal stage \cite{yang2018ipod, shi2019pointrcnn, qi2018frustum}. 

In this paper, we revisit the design of object detection systems in the context of 3D LiDAR data, and propose a new framework which better matches the data modality and the demands of SDC perception.

We start by recognizing that 3D region proposals are fundamentally distinct. Every reflected point must belong to an object or surface. In this setting, we demonstrate that efficient sampling schemes on point clouds -- with zero learned parameters -- are sufficient for generating region proposals. In addition to being computationally inexpensive, sampling has the advantage of implicitly exploiting the sparsity of the data by matching the data distribution of the scene. 

Departing from the trend of increasing use of global context, we process each proposed region completely independently. This independence, and non-learned proposal mechanism also allows us to inject priors into the proposal process, which we show the value of by leveraging temporal context in the form of seeding sampling with the previous frames detections. Finally, we entirely avoid any discretization procedure and instead featurize the native point cloud \cite{qi2017pointnet, qi2017pointnet++}
in order to classify objects and regress bounding box locations \cite{faster_rcnn, ren2015faster}. 

The resulting detector is as accurate as the state of the art at lower inference cost, and more accurate at similar inference costs.
In addition, these design decisions result in several key benefits.
First, the model does not waste computation on empty regions because the proposal method naturally exploits point cloud sparsity.
Second, one can dynamically vary the number of proposals and the number of points per proposal at inference time since the model operates locally. This feature permits a single trained model to operate at different computational budgets.
Third, the model can easily leverage contextual information (HD maps, temporal context) to target computation. For example, detection outputs from preceding frames can be used to inform the current frame's sampling locations.

In summary, our main contributions are as follows: 

\begin{compactitem}
    \item Introduce a flexible, local point-based object detector geared towards SDC perception. In the process we demonstrate that cheap proposals on point clouds, paired with a point-based network, results in a system that is competitive with state-of-the-art performance on self-driving car benchmarks.
    \item Demonstrate the computational-flexibility of our model through showing how a single model designed in this fashion may adapt its inference cost. For instance, a single trained pedestrian model may exceed the predictive performance of a baseline convolutional model by $\sim48\%$ at similar computational demand; or, the same model without retraining may achieve similar predictive performance but with $\sim20\%$ of the computational demand.
    \item Demonstrate the ability of the model to selectively target specific locations of interest. We show how temporal context (using only the \emph{outputs} of previous frames) can be used with the model to improve detection mAP scores by $\sim40\%$.
\end{compactitem}
\begin{figure*}[h!]
\centering
\includegraphics[width=0.9\linewidth]{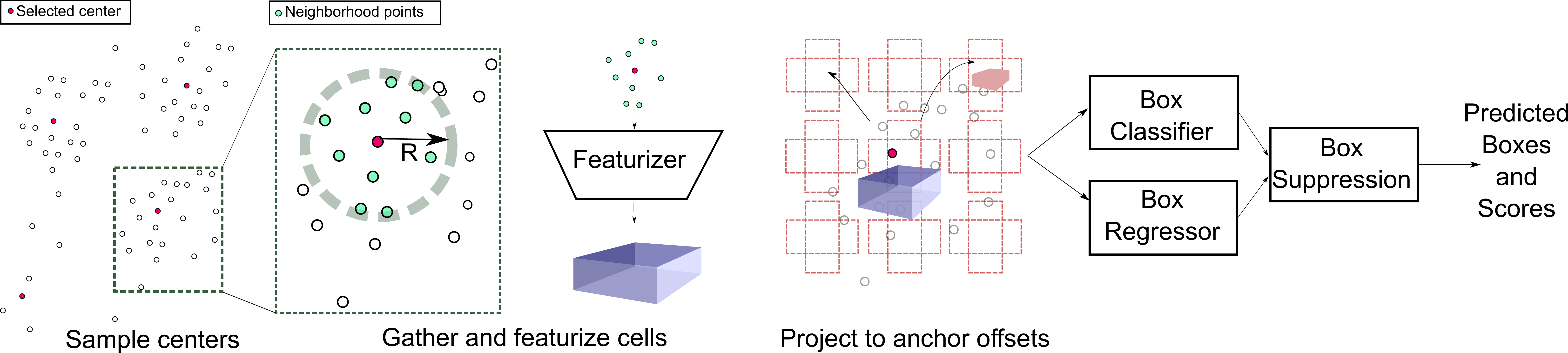}
\caption{\textbf{\name\ overview}.}
\label{fig:overview}
\end{figure*}

\section{Background}

\subsection{Object detection in images}

Early object detection systems consist of two stages: first, to propose candidate detection locations, and next, to discriminate whether a given proposal is an object of interest \cite{felzenszwalb2010object, dean2013fast,sermanet2013pedestrian,uijlings2013selective}. The advent of convolutional neural networks (CNN) \cite{krizhevsky2012imagenet,lecun2015deep},
showed that a CNN-based featurization may provide superior proposals as well as improve the second discriminative stage \cite{girshick2014rich,ren2015faster,girshick2015fast}.
Modern CNN-based detection systems maximize prediction performance by densely sampling an image for all possible object locations. This requires a computationally-heavy first stage featurization to provide high quality bounding box proposals \cite{girshick2014rich, girshick2015fast}. In addition, the second stage of an object detector will need to be run on each proposal within a {\it single} image.
These heavy computational demands are prohibitive in constrained environments (e.g. mobile and edge devices) such that speed versus accuracy trade-offs must be considered \cite{huang2017speed}.

To address these concerns, recent work has focused on {\it one stage} object detectors that attempt predict bounding box locations and object identity in a single inference operation of a CNN \cite{liu2016ssd,redmon2016you}. Although single-stage systems provide faster inference, these systems generally exhibit worse predictive performance than two-stage systems \cite{huang2017speed}. That said, recent advances in redesigning the loss functions have mitigated this disadvantage significantly
\cite{lin2017focal,zhou2019bottom,law2018cornernet}.
The speed and reduced complexity advantages associated with a one stage model do however come with an associated cost: by basing the inference procedure on convolutions which densely sample an image, the resulting model must treat all spatial locations equally. In the context of self-driving cars, this design decision hampers the ability to adapt computation to the current scene or latency requirements.

\subsection{Point cloud featurization}

Raw data arising from many types of sensors come in the form of point cloud data (e.g. LIDAR, RGB-D). A point cloud consists of a set of $N$ 3-D points $\{\vec{x}_i\}$ indexed by $i$ which may contain an associated feature vector $\{\vec{f}_i\}$. The set of points are unordered and may be of arbitrary size depending on the number of reflections identified by a sensor on a single scan. Ideally, learned representations for point clouds aim to be permutation invariant with respect to $i$ and agnostic to the number of points $N$ in a given example \cite{qi2017pointnet, qi2017pointnet++}. On-going efforts have attempted to design models that operate directly on point cloud data, some of which are derived to mimic convolutions \cite{wang2018deep,zaheer2017deep}.

\subsection{Object detection with point clouds}
\label{sec:point-cloud-detection}

Object detection in point clouds has started with porting ideas from the image-based object detection literature. By voxelizing a point cloud (i.e. identifying a grid location for individual points $\vec{x}_i$) into a series of stacked image slices describing occupancy, one may employ traditional CNN techniques for object detection on the resulting images or voxel grids  \cite{yang2018pixor,zhou2018voxelnet,yan2018second,luo2018fast,yang2018hdnet,yin2019multiview,meyer2019lasernet}.

VoxelNet partitions 3-D space and encodes LiDAR points within each partition with a point cloud featurization \cite{zhou2018voxelnet}. The result is a fixed-size feature map, on which a conventional CNN-based object detection architecture may be applied.
Likewise, PointPillars \cite{lang2018pointpillars} proposes an object detector that employs a point cloud featurization, providing input into a grid-based feature representation for use in a feature pyramid network \cite{lin2016feature}; the resulting per-pillar features are combined with anchors for every pillar to perform joint classification and regression. 
The resulting network achieves a high level of predictive performance with minimal computational cost on small scenes, but its fixed grid increases in cost notably on larger scenes and cannot adapt to each scene's unique data distribution.

LaserNet \cite{meyer2019lasernet} opts to work on the Range Image representation of the LiDAR instead of a point cloud or a voxelized view. The range image data takes on a dense perspective view that LaserNet applies convolutions to. While this method has the advantage of working on a dense representation, it also faces challenges (similar to camera images) of having a perspective effect on the scale objects. Objects in a range image may have a large variance in scale.

In the vein of two stage detection systems, PointRCNN \cite{shi2019pointrcnn} employs a point cloud featurizer \cite{qi2017pointnet++} to make proposals via an expensive per-point segmentation network into foreground and background. Subsequently, a second stage operates on cropped featurizations to perform the final classification and localization. Other works propose bounding boxes through a computationally intensive, learned proposal system operating on paired camera images~\cite{yang2018ipod,qi2018frustum}, with the goal of improving predictive performance by leveraging a camera image to seed a proposal system to maximize recall.

\section{Methods}
\label{sec:methods-approach}
Our goal is to construct a detector that better aligns with the requirements of a SDC perception system, taking advantage of the sparsity of the data, allowing us to target where to spend computation, and operating on the native data. 
To address these goals, we propose a sparse targeted object detector, termed {\it \name} (Figure \ref{fig:overview}):
Given a sparse sampling of locations in the point cloud, the model extracts a small (random) subset of neighboring points. The model featurizes the point cloud \cite{qi2017pointnet}, classifies the region, and regresses bounding box parameters.
The object location is predicted {\it relative} to the selected location and only uses local information. This setup ensures that each spatial location may be processed by the detector independently.

The structure of the proposed system confers two advantages. First, inference on each cell proposal occurs completely independently, enabling computation of each location to be parallelized to decrease inference latency. Second, contextual information information \cite{bojarski2016end,yang2018hdnet} may be used to inform importance of each proposal. 

The rest of this section describes the architecture of \name\ in more detail.

\subsection{Center selection}
\label{sec:center-selection}

We propose using an inexpensive, data-dependent algorithm to generate proposals from LiDAR with high recall. In contrast to prior work \cite{yang2018pixor,lang2018pointpillars,yan2018second}, we do not base proposals on fixed grid locations, but instead generate proposals to respect the observed data distribution in a scene.
\begin{figure}
\centering
\includegraphics[width=0.45\linewidth]{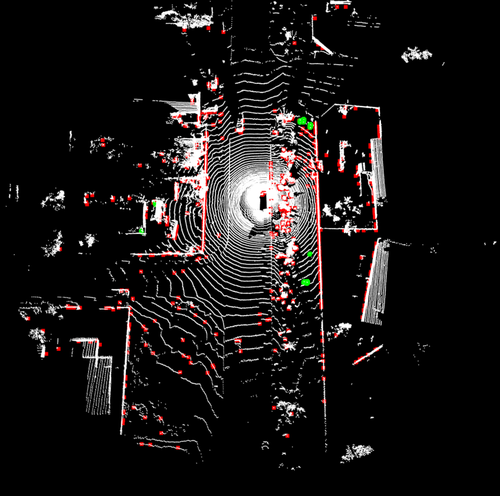}
\includegraphics[width=0.45\linewidth]{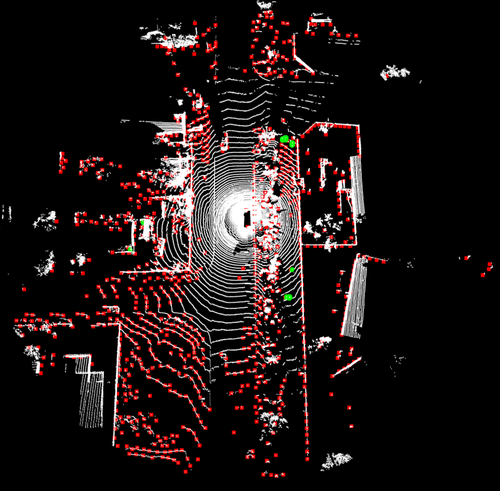}
\captionof{figure}{Example of random uniform sampling (left) and farthest point sampling (right) with the \textit{same number of samples}. Red indicate selected centers. Green indicate pedestrians. Note that random uniform sampling biases towards high density regions, while farthest point sampling evenly covers the space. Neither place any proposals in empty space.}
\label{fig:center-selection}
\vspace{-5pt}
\end{figure}

Concretely, we sample $N$ points from the point cloud, and use their $(x, y)$ coordinates as proposals. To avoid sampling regions on the ground, we follow previous works \cite{yan2018second, lang2018pointpillars} and only allow sampling of points between a certain $z$-dimension range. For KITTI \cite{geiger2013vision} this is $z \in [-1.35, \inf]$, and for the \datasetname\ \cite{waymo_open_dataset} we calculate the $10^{th}$ and $90^{th}$ percentile of the center z location of all objects. Note that these points are only excluded for sampling, and will be present in later stages. 

In this work, we explore three sampling algorithms: random uniform sampling, farthest point sampling (FPS), and a hybrid approach of seeding FPS with preceding frame detections (Figure \ref{fig:center-selection}, Section \ref{temporal_proposals}). Random uniform sampling provides a simple and effective baseline because the sampling method biases towards densely populated regions of space. In contrast, farthest point sampling (FPS) selects individual points sequentially such that the next point selected is maximally far away from all previous points selected, maximizing the spatial coverage across the point cloud. Finally, in Section \ref{temporal_proposals}, we show how to leverage the previous frame detection outputs as seed locations for FPS. We show that this is a light-weight and effective way to leverage temporal information.

\subsection{Featurizing local point clouds}

After obtaining a proposal location, we featurize the local point cloud around the proposal. We randomly select $K$ points within a radius of $R$ meters of each proposal center.  In our experiments, $K$ is typically between 32 to 1024, and $R$ is 2-3 meters.  All local points are re-centered to an origin for each proposal.  LiDAR features associated with each point are also used as part of the input. 
We experimented with several architectures for featurizations of native point cloud data \cite{qi2017pointnet++, wu2018pointconv} but most closely followed \cite{xu2018powerful}. The resulting architecture is agnostic to the number of points provided as input \cite{qi2017pointnet++, wu2018pointconv, xu2018powerful}.

\begin{figure}
\centering
\includegraphics[width=0.5\linewidth]{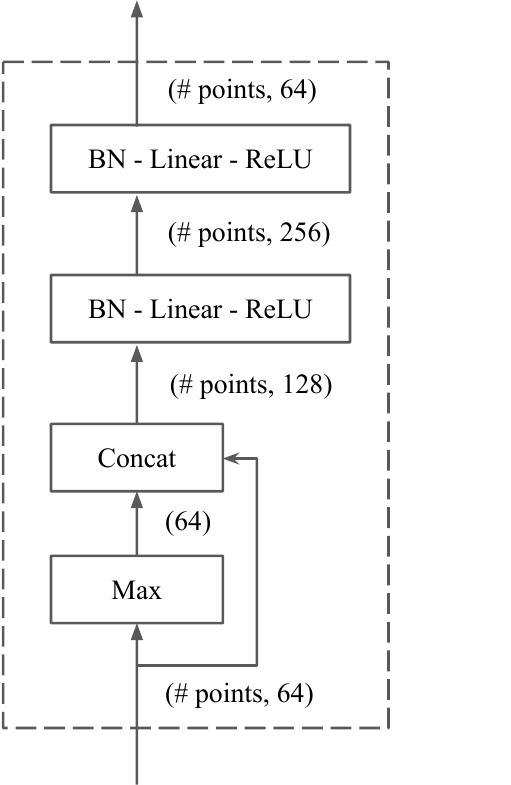}
\caption{\textbf{\name\ Block}.
We annotate edges with tensor dimensions for clarity:  (\# points, 64) represents a point cloud with \# points, where each point has an associated 64-dimensional feature.
}
\label{fig:starnet_block}
\end{figure}

\begin{figure}[h]
\centering
\includegraphics[width=0.8\linewidth]{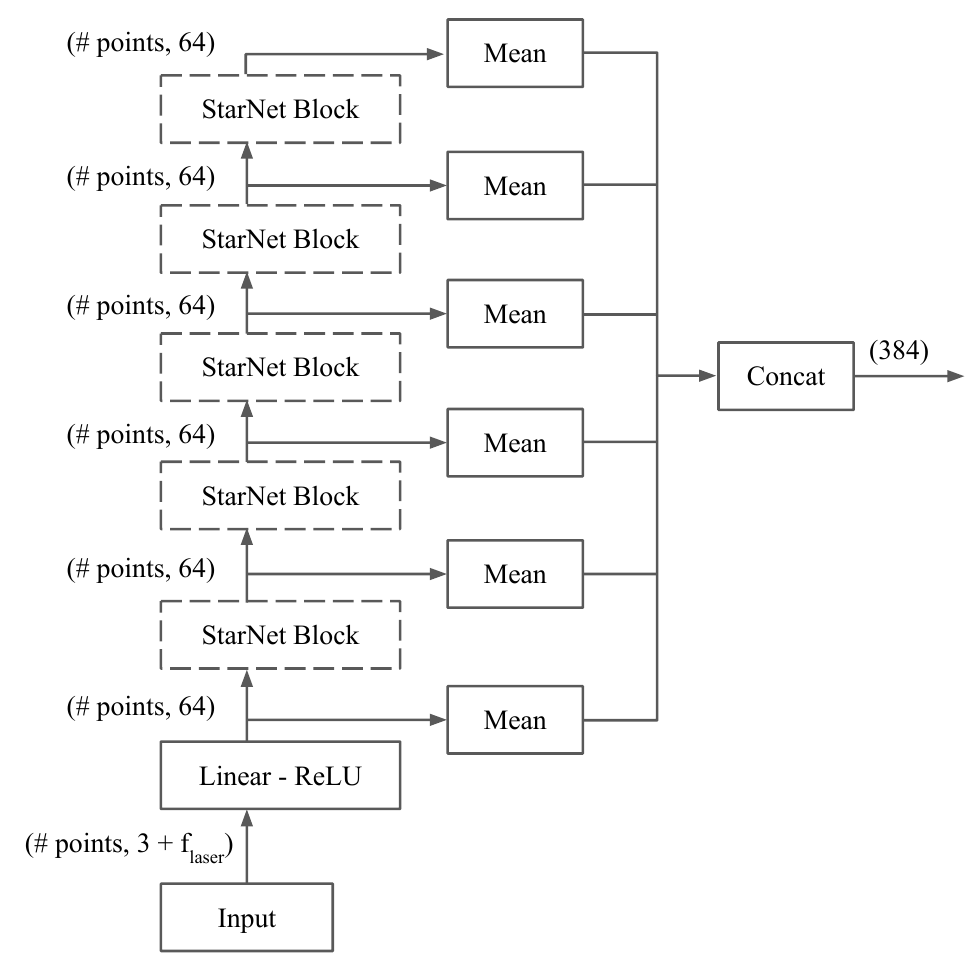}
\caption{\textbf{\name\ point cloud featurizer.} StarNet blocks are stacked, where each block's output is read out using mean aggregation.  The readouts are concatenated together to form the featurization for the point cloud. }
\label{fig:starnet_point_featurzier}
\vspace{-10pt}
\end{figure}

StarNet blocks (Figure \ref{fig:starnet_block}) take as input a set of points, where each point has an associated feature vector.  Each block first computes  aggregate statistics (max) across the point cloud.  Next, the global statistics are concatenated back to each point's feature.  Finally, two fully-connected layers are applied, each composed of batch normalization (BN), linear projection, and ReLU activation. StarNet Blocks are stacked to form a 5-layer featurizer (Figure \ref{fig:starnet_point_featurzier}) that outputs a 384-dimensional feature.

\begin{figure*}[t!]
\centering
\includegraphics[width=0.4\linewidth]{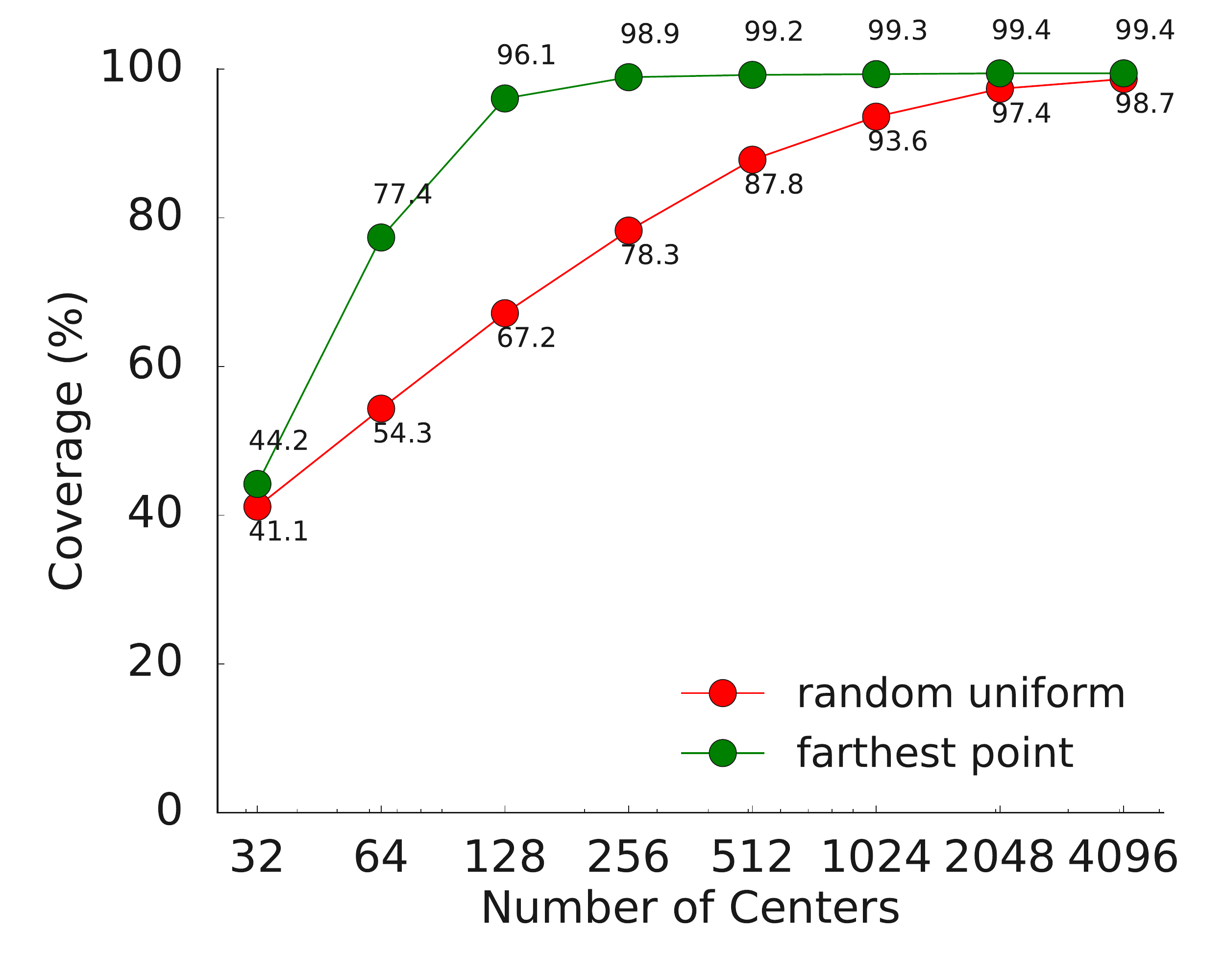}
\includegraphics[width=0.4\linewidth]{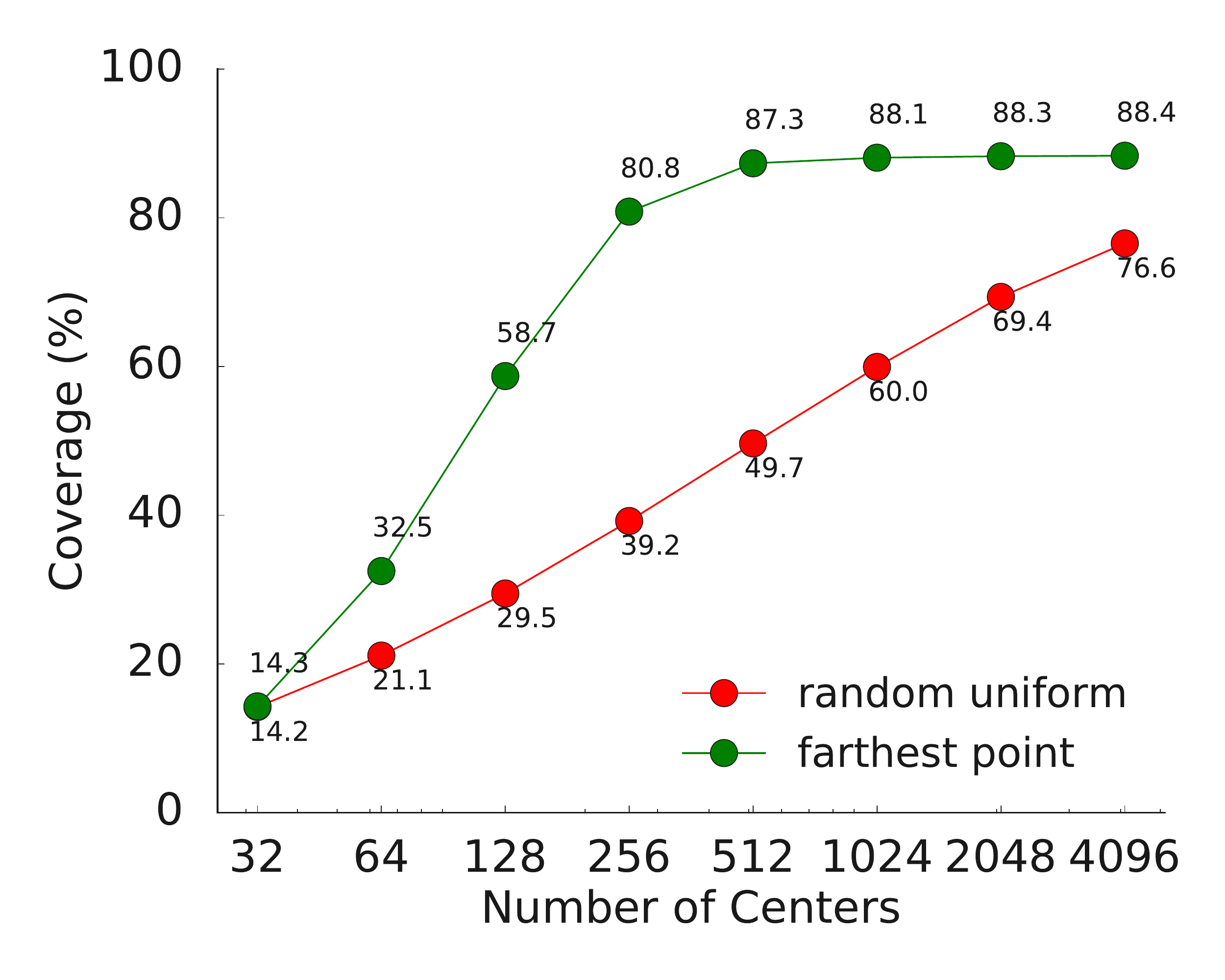}
\caption{{\bf Simple sampling procedures have good coverage over ground truth bounding boxes.} The coverage of proposals for cars and vehicles is plotted against the number of samples on KITTI (left) and \datasetname\ (right). Error bars (not shown) range from 0.5\%-3.0\%. See text for details.}
\vspace{-10pt}
\label{fig:recall-for-proposals}
\end{figure*}

The StarNet point featurizer (Figure \ref{fig:starnet_point_featurzier}) stacks multiple StarNet blocks, following ideas from graph neural networks \cite{xu2018powerful}.   We experimented with different choices network architectures and found that using max aggregation, concatenate combination, and mean readout performed well.  By design, the same trained network can be used with varying number of input points, giving it a large degree of flexibility.  

\subsection{Constructing final predictions from bounding box proposals.}

For each cell center, a grid of $G \times G$ total anchor offsets are placed relative to each cell center, where each offset can employ different rotations or anchor dimension priors.  We emphasize that unlike single-stage detectors \cite{lang2018pointpillars, yan2018second}, the anchor grid placement is data-dependent since it is based on the proposals.

For each grid offset, we compute a $D$ dimensional feature vector using a learned linear projection from the cell's 384-dimensional feature; each offset has a different projection. The $D$ dimensional feature is shared across the rotations and dimensions at the grid offset. From this feature, we predict classification and regression logits. The bounding box regression logits predict $\delta x, \delta y, \delta z$ corresponding to residuals of the location of the anchor bounding box; $\delta h, \delta w, \delta l$ corresponding to residuals on height, width and length; and a residual on the heading orientation $\delta \theta$. 

We use a smoothed-L1 loss on each predicted variate \cite{yan2018second,lang2018pointpillars,yin2019multiview}. For the rotation loss, 
We use a direction invariant loss $\textrm{SmoothL1}(sine(\delta \theta - {\delta \theta}_{\textrm{gt}}))$ for all experiments, except for models where we report heading accuracy weighted average precision (mAPH). For direction aware models, we use $\textrm{SmoothL1}(WrapAngle(\delta \theta - {\delta \theta}_{\textrm{gt}}))$, where WrapAngle ensures the angular difference is between $-\pi$ to $\pi$. The classification logits are trained with a focal cross-entropy loss on the class label \cite{lin2017focal}. 

Ground truth labels are assigned to anchors based on their intersection-over-union (IoU) overlap~\cite{yan2018second, lang2018pointpillars}. We compute the IoU for each anchor and ground truth label and assign labels to foreground if $IoU > 0.6$ or background if $IoU < 0.45$. Anchors with IoU matches between the two thresholds are ignored. We also perform force-matching if an object is not assigned as foreground to any anchor: we assign the object as foreground to its highest matching anchor if (a) the highest matching anchor is not assigned to foreground of any object and (b) the IoU with the matching anchor is greater than zero. Final predictions use an oriented, multi-class non-maximal suppression (NMS) \cite{girshick2014rich}.

\begin{figure}[h!]
\centering
\includegraphics[width=0.9\linewidth]{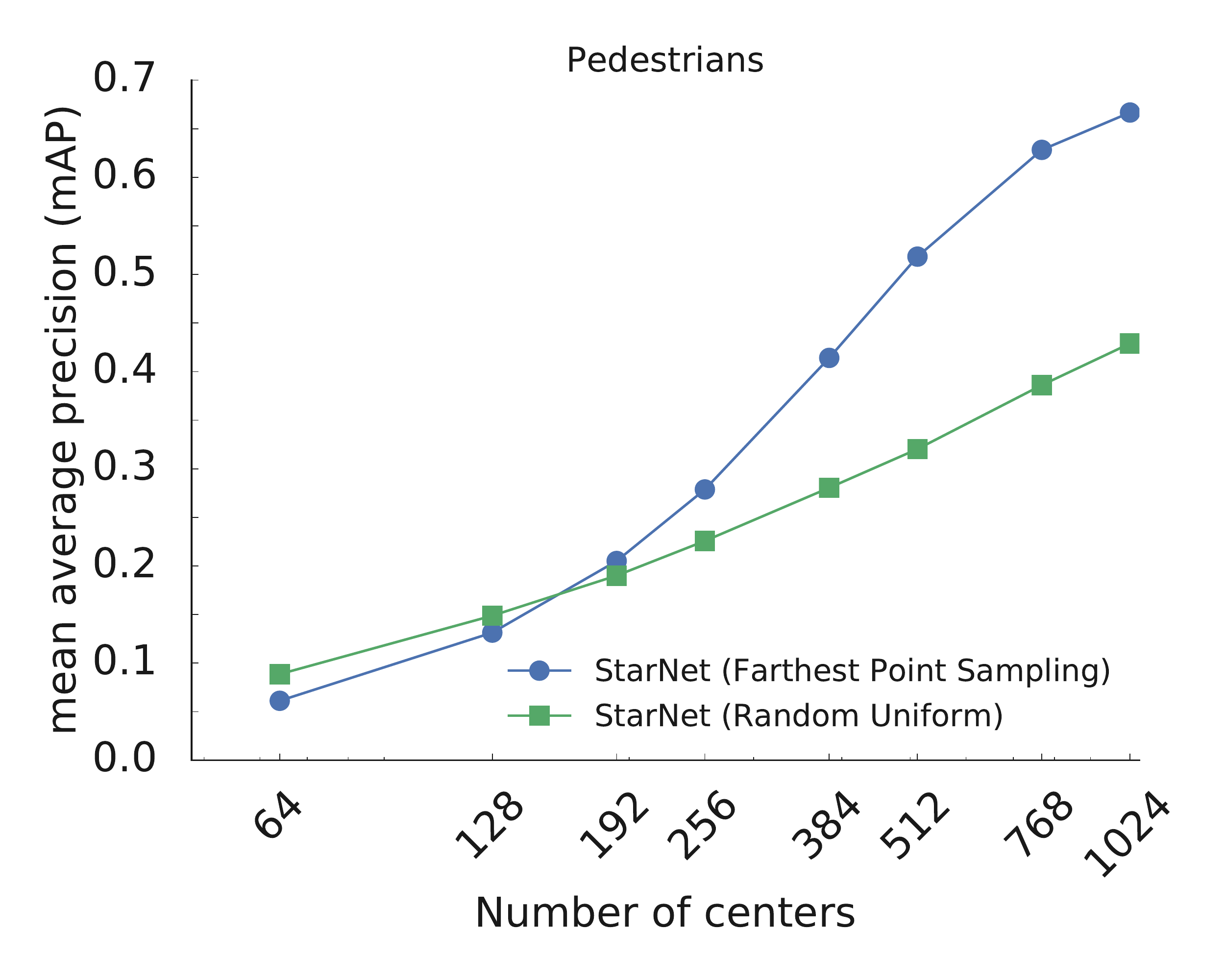}
\caption{\textbf{Adaptive computation with a single trained model}. \datasetname\ \textit{Validation set} mAP on pedestrians of a single \name\ model trained with 1024 proposals, evaluated with 64 to 1024 proposals.}
\vspace{-10pt}
\end{figure}
\begin{table*}[t]
\footnotesize
\begin{center}
\bgroup
\def\arraystretch{1.4}%
\begin{tabular}{| c || c | c | c || c | c | c || c | c | c |}
\hline
\multirow{2}{20mm}{\it 3D detection} 	& \multicolumn{3}{|c||}{Car}		& \multicolumn{3}{|c||}{Pedestrian}		& \multicolumn{3}{|c|}{Cyclist}	\\ \cline{2-10}
					& Easy		& Mod.		& Hard		& Easy		& Mod.		& Hard		& Easy		& Mod.		& Hard	\\ \hline 
VoxelNet~\cite{zhou2018voxelnet}			& 77.47		& 65.11  		& 57.73		& 39.48		& 33.69			& 31.5		& 61.22		& 48.36		& 44.37	\\
SECOND~\cite{yan2018second}			& 83.13 	& 73.66	 	& 66.20		& 51.07		& 42.56			& 37.29		& 70.51		& 53.85 		& 46.90 	\\
PointPillars \cite{lang2018pointpillars}			& 79.05		& 74.99	& 68.30	& 52.08	& 43.53			& 41.49		& 75.78 		& 59.07		& 52.92	\\
\hline
\name	& 		81.63		& 73.99	& 67.07  &   48.58 		& 41.25		& 39.66 &	73.14 	& 58.29			& 		52.58 	\\

\hline
\end{tabular}
\egroup
\vspace{0.4cm}

\bgroup
\def\arraystretch{1.4}%

\egroup
\end{center}
\vspace{-20pt}
\caption{Results on the KITTI {\it test} object detection benchmark for object detection systems using 3-D evaluation. All detection results and comparisons based only on LIDAR data. mAP calculated with an IOU of 0.7, 0.5 and 0.5 for vehicles, cyclists and pedestrians, respectively.}
\vspace{-10pt}
\label{table:kitti-test}
\end{table*}

\section{Results}
We present results on the KITTI object detection benchmark \cite{geiger2013vision} and the \datasetname\ \cite{waymo_open_dataset}. We train models using the Adam \cite{kingma2014adam} optimizer with an exponentially-decaying learning rate schedule starting at 1e-3 and decaying over 650 epochs for KITTI, and 75 epochs for the \datasetname. We perform some hyper-parameter tuning on the validation set and perform final evaluations on the corresponding test datasets. Full hyperparameters can be found in our already open-sourced code (\ifcvprfinal \url{http://github.com/tensorflow/lingvo}\else \url{http://github.com/anonymized}\fi). 

\subsection{Sampling strategies for point cloud detections}
We first investigate sampling strategies for center selection,  evaluating on KITTI and \datasetname. We explore two strategies for naively sampling point clouds: random sampling and farthest point sampling (Section \ref{sec:center-selection}). We observe that random sampling samples many centers in dense locations, whereas farthest point sampling maximizes spatial coverage of the scene. 

To quantify the efficacy of each proposal method, we measure the coverage as a function of the number of proposals. Coverage is defined as the fraction of annotated objects with 5+ points that have IoU $> 0.5$ with the our sampled anchor boxes. Figure \ref{fig:recall-for-proposals} plots the coverage for each method for a fixed IOU of 0.5 for cars in KITTI \cite{geiger2013vision} and the \datasetname\ \cite{waymo_open_dataset}. All methods achieve monotonically higher coverage with greater number of proposals with coverage on KITTI exceeding 98\% within 256 samples. 
Because random sampling is heavily biased to regions which contain many points, there is a tendency to oversample large objects and undersample regions containing few points. Farthest point sampling (FPS) uniformly distributes samples across the spatial extent of the point cloud data (see Methods). We observe that FPS provides uniformly better coverage across a fixed number of proposals and we employ this technique for the rest of the work.

\subsection{KITTI Dataset}
\label{sec:two-stage-kitti}

When evaluating \name~on the KITTI dataset, we found that data augmentation important to obtain good performance. We employed standard data augmentations for point clouds and bounding box labels \cite{yang2018pixor, zhou2018voxelnet, yan2018second, luo2018fast, yang2018hdnet, lang2018pointpillars}. We found that the gains in predictive performance due to data augmentation (up to +18.0, +16.9 and +30.5 mAP on car, pedestrian and cyclist respectively) were substantially larger than gains in performance observed across advances in detection architectures. Additionally, we found checkpoint selection to be extremely important due to the small size of the dataset, and submission filtering (e.g. removing detections where the 2D projected height of our 3D bounding box predictions were smaller than 25 pixels so they are not erroneously labeled as false positives) unique challenges to the KITTI benchmark. 

We take our best system for 3-D object detection with the same data augmentations and compare the efficacy of this model to previously reported state-of-the-art systems that only operate on point cloud data \cite{zhou2018voxelnet, yan2018second, lang2018pointpillars, yang2018hdnet}. Table \ref{table:kitti-test} reports the 3-D
detection results on the KITTI {\it test} server. \name\ provides
competitive mAP scores on car, pedestrian and cyclist to other state-of-the-art methods, exceeding subsets of each category strata. 

We found that decisions apart from model design play a significant role in KITTI test set performance: this included data augmentation, checkpoint selection, post process filtering, among others. Since we are interested in determining the efficacy of our modeling approach, we focus the majority of our following experiments on the larger \datasetname, which is annotated with high quality labels.

\begin{table*}[t]
\begin{center}

\begin{tabular}{cc}

\begin{minipage}{.5\linewidth}

\begin{tabular}{l|c||c|c}
Model & Directional? & mAP & mAPH \\
\hline
PointPillars* \cite{lang2018pointpillars} & $\checkmark$ & 60.0 & 47.3 \\
StarNet & -- & \textbf{70.1} & 35.6 \\
StarNet & $\checkmark$ & 67.8 & \textbf{59.9} \\
\hline
StarNet & -- & \textbf{72.1} & 37.0 \\
(with temporal context)  & & & 
\end{tabular}

\end{minipage} &

\begin{minipage}{.5\linewidth}

\begin{tabular}{l|c||c|c}
Model & Directional? & mAP & mAPH \\
\hline
PointPillars* \cite{lang2018pointpillars} & $\checkmark$ & 62.2 & \textbf{61.7} \\
StarNet & -- & \textbf{64.7} & 45.5 \\
StarNet & $\checkmark$ & 61.5 & 61.0 \\
\hline
StarNet & -- & \textbf{65.0} & 45.6 \\
(with temporal context)  & & & 
\end{tabular}
\end{minipage}  \\

(a) Pedestrian Detection & (b) Vehicle Detection

\end{tabular}

\end{center}
\caption{Waymo Open Dataset \textit{Test set} results on the \textit{LEVEL\_1} category ($\geq$ 5 points) for StarNet versus a reimplemented PointPillars \cite{lang2018pointpillars} baseline model. Directional indicates a directional aware heading loss was used. Temporal context models use the top 512 highest confidence detected locations from the \textit{previous} frame as anchor centers, with the remaining 512 centers drawn using Farthest Point Sampling.}
\label{waymo_test_results}
\end{table*}
\begin{table}[]
\begin{tabular}{l | c | c }
Model             & Pedestrian mAP & Vehicle mAP \\
\hline
PointPillars* \cite{lang2018pointpillars} & 62.1           & 57.2        \\
Multi View Fusion \cite{yin2019multiview} & 65.3          & \textbf{62.9}      \\
\hline
StarNet           & \textbf{66.8}  & 53.7       
\end{tabular}
\caption{\datasetname\ \textit{Validation set} results of LEVEL\_1 difficulty for comparison with Multi-View Fusion \cite{yin2019multiview} and a reimplemented PointPillars \cite{lang2018pointpillars}.}
\label{waymo_val_table}
\vspace{-10pt}
\end{table}

\subsection{Waymo Open Dataset}
\label{sec:real-world}

We now focus on the performance of \name\ on the \datasetname\ \cite{waymo_open_dataset}, which is substantially larger and exhibits tremendous diversity.

\begin{figure*}[h]
\centering
\includegraphics[width=0.4\linewidth]{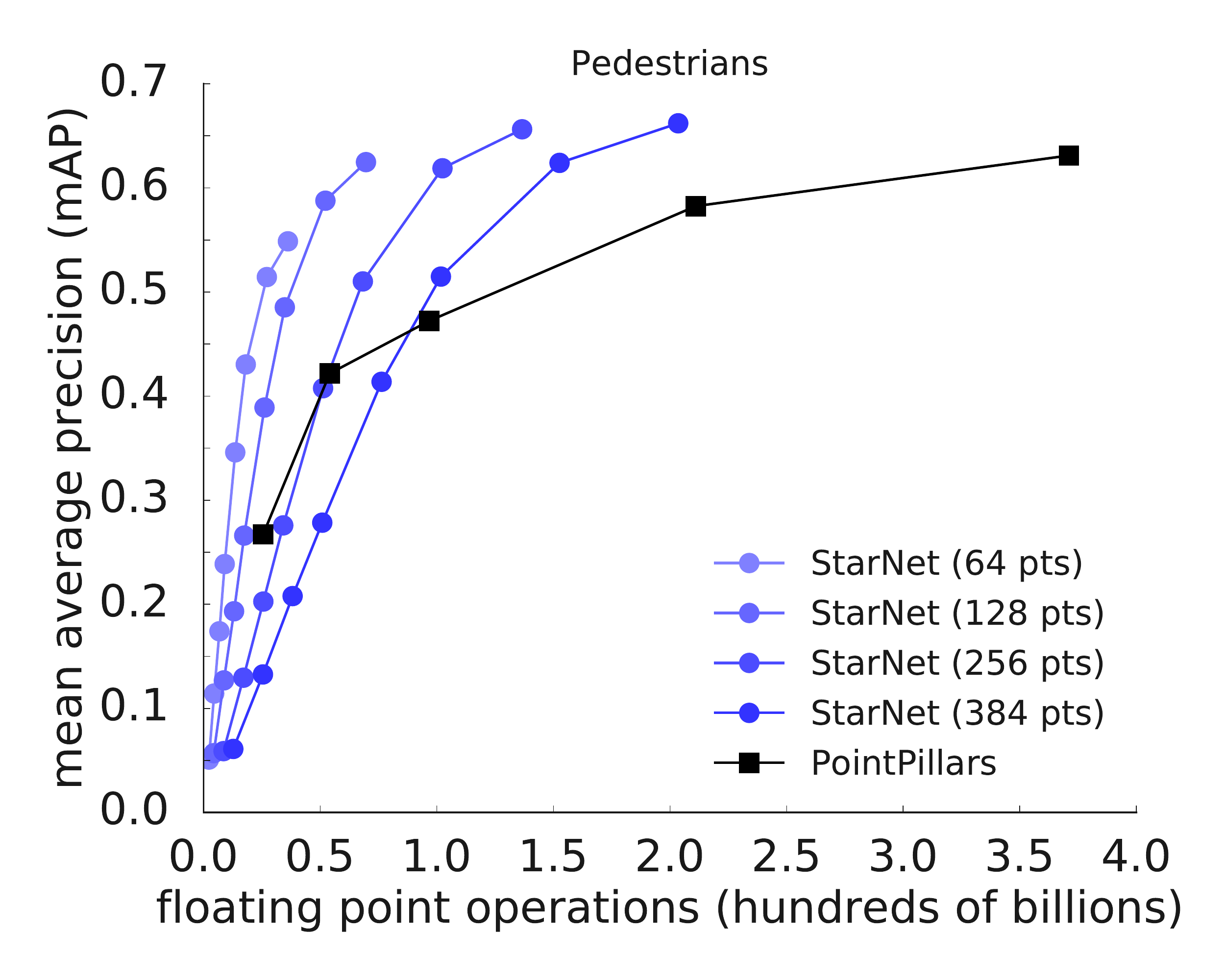}
\includegraphics[width=0.4\linewidth]{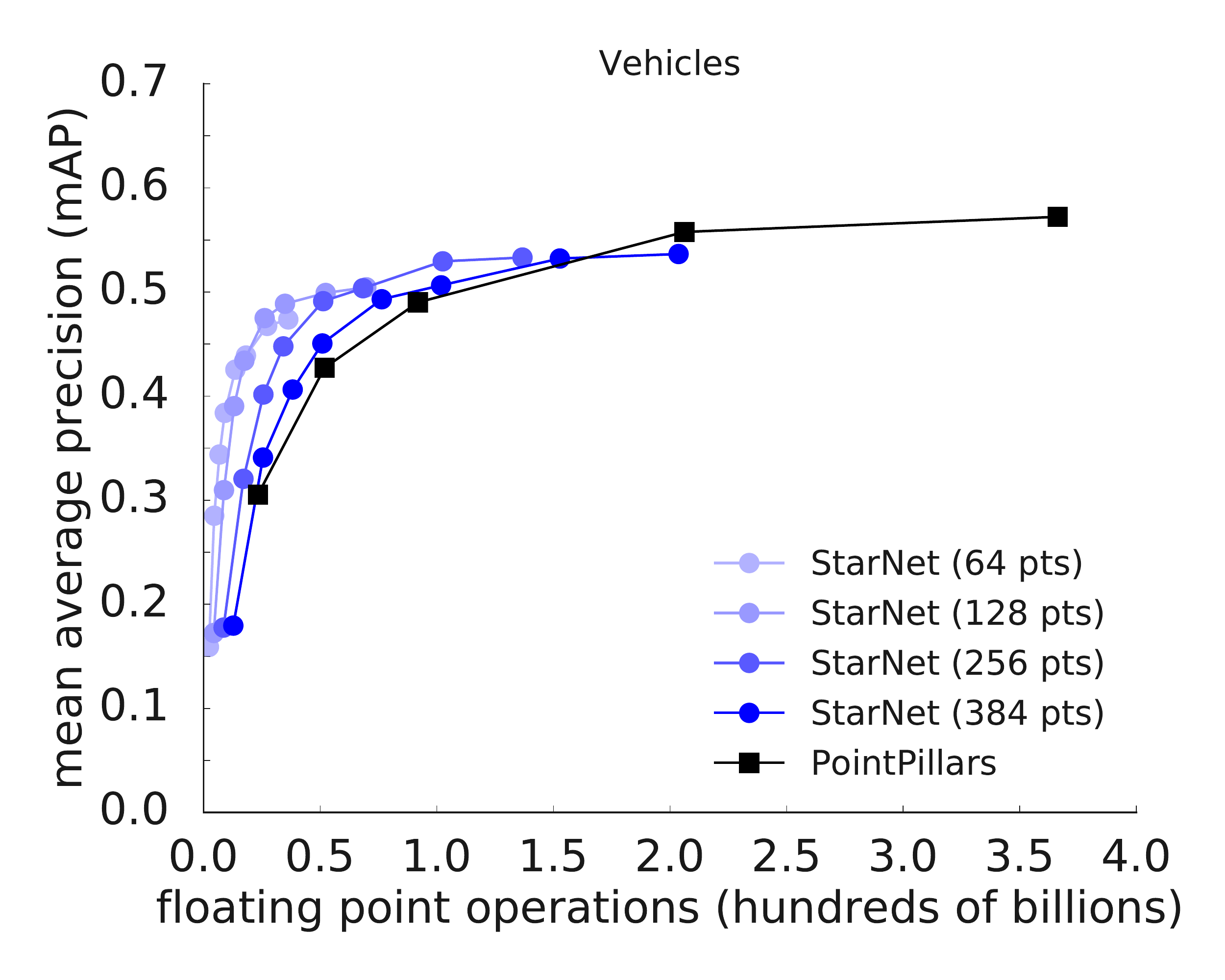}
\caption{\textbf{Flexible computational cost of detection} for (left) pedestrians and (right) vehicles.
Across 5 separately-trained PointPillars models \cite{lang2018pointpillars}, computational cost grows quadratically with increased spatial resolution for the LiDAR pseudo-image.
All curves for \name\ arise from a {\it single} set of saved model weights.
Each curve traces out \name\ accuracy on the \textit{Validation set} for a fixed number of point cloud points.
Points along on a single curve indicate 64 to 1024 selected centers.}
\label{fig:waymo-detection}
\end{figure*}

To demonstrate the relative merits of \name, we trained models on pedestrians and vehicles and compared the relative performance of each model to a family of baseline models. Data augmentation was not used in these experiments. We employed PointPillars as a baseline model\footnote{We note that our custom implementation of PointPillars achieves 74.5, 57.1, and 59.0 mAP for for cars, pedestrians, and cyclists, respectively on KITTI validation at moderate difficulty. This is slightly lower than \cite{lang2018pointpillars}.}, training 5 different grid resolutions of this model for each class and validated accuracy on all annotated bounding boxes with 5+ LiDAR points. Each version employs a different input spatial resolution for the pseudo-image (128, 192, 256, 384, and 512 pixel spatial grids), with 16K to 32K non-zero featurized locations (pillars). In slight deviation from the original PointPillars paper \cite{lang2018pointpillars}, we use an output stride of 1 for both vehicles and pedestrian models, as it exhibits substantially higher performance. We hypothesize that a single-stage object detector would exhibit trade-offs in detecting small objects based on the resolution of the image projection. Indeed, we observe in Figure \ref{fig:waymo-detection} (black points) that higher spatial resolutions achieve higher precision for pedestrians and vehicles, but with a computational cost that grows quadratically.

We also examined the performance of a {\it single} \name\ model across
two strategies for altering computational demand: varying the number of proposals, and varying the number of points supplied to the model per proposed region.
Each blue curve in Figure \ref{fig:waymo-detection} traces out the computational cost versus predictive performance for a given number of points per region, while varying the number of proposals from 64 to 1024. Many of these points lie above the baseline model indicating that \name\ provides favorable performance.
In particular, the same pedestrian detection model (e.g., \name-128 with 1024 centers) may achieve $\sim48\%$ relative gain in predictive performance for a similar computational budget as the baseline pillars model ($\sim100GFlops$); or, the same model achieves similar predictive performance as the most accurate Pillars model but with $\sim20\%$ of the computational budget.
We emphasize that all \name~points arise from a {\it single} trained model, showing how to use a single trained \name\ in a flexible manner through manipulations at inference time.

Finally, we took the highest performing \name\ and PointPillars \cite{lang2018pointpillars} models from Figure \ref{fig:waymo-detection}, and evaluated them on the \datasetname \cite{waymo_open_dataset} \textit{Test set}. These results are summarized in Table \ref{waymo_test_results}, with full numbers including range based breakdowns available in Appendix A. \name\ is competitive on Vehicle detection to our PointPillars \cite{lang2018pointpillars} baseline, and significantly outperforms it for Pedestrians. If a directional loss is used, we outperform PointPillars by 7.8 mAP and 12.6 mAPH, and 10.1 mAP if a directionless loss is used. Note that forcing the network to learn directionality slightly hinders mAP. Additionally, using temporal context, detailed next, further improves performance.  
We also compare to Multi-View Fusion \cite{yin2019multiview} in Table~\ref{waymo_val_table}, showing validation set results as the Multi-View Fusion method does not yet report test set numbers.

\subsection{Targeting computation with temporal context}
\label{temporal_proposals}

One design benefit of \name\ is the ability to target computation. We now show how using the outputs of the previous time-step can significantly improve mAP over a single frame, while keeping the computation cost unchanged.

\begin{figure}
\centering
\includegraphics[width=0.85\linewidth]{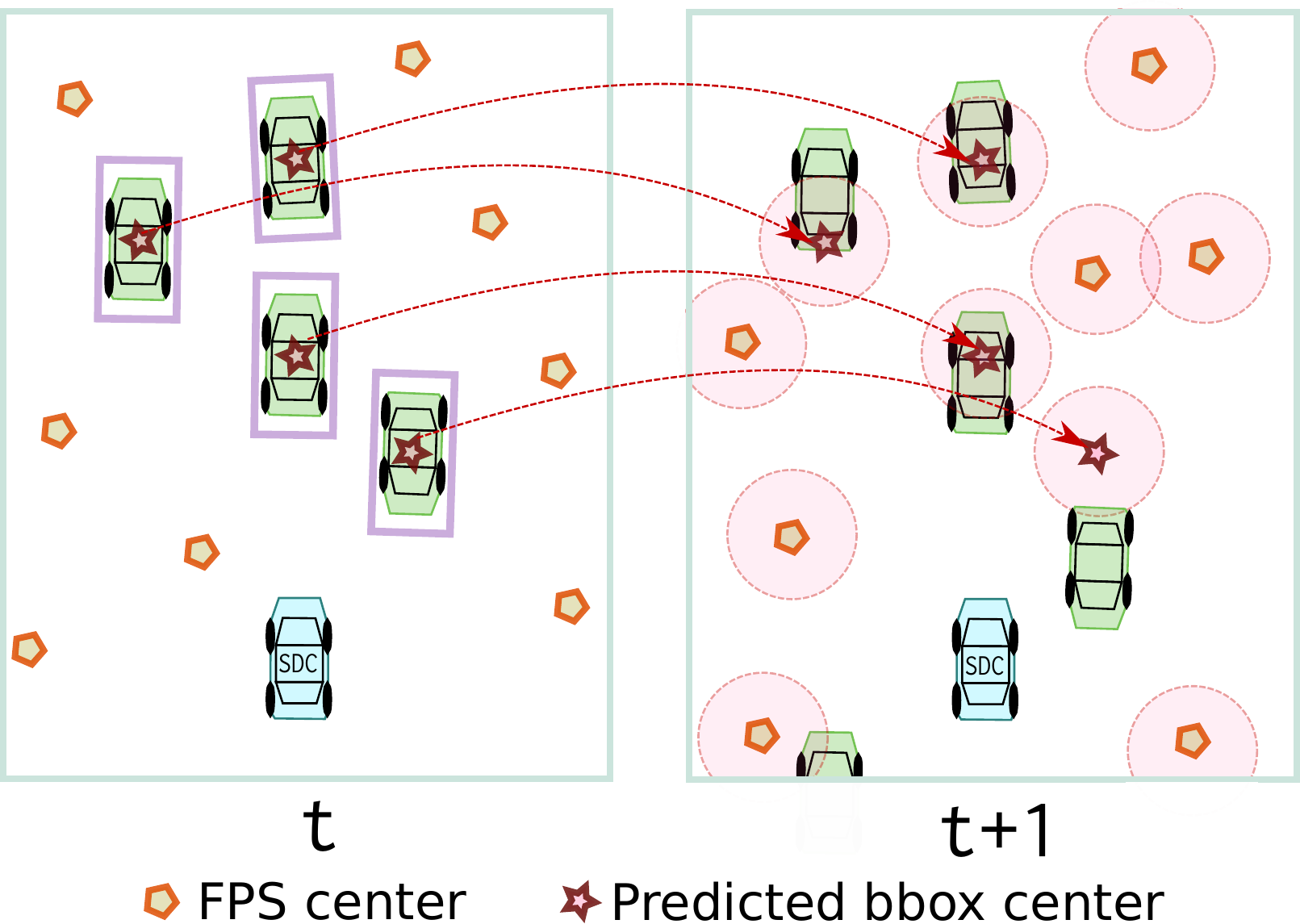}
\caption{\textbf{Leveraging previous proposals}. Using the highest confidence predicted centers from the previous frames can help improve detection mAP in the next frame.
}
\label{fig:starnet_seed}
\vspace{-10pt}
\end{figure}

Intuitively, high-confidence bounding box proposals output on previous time-steps in 3D are a good prior on the location of objects in the current frame since objects have limited ranges of motion. Hence, one natural approach is to leverage these priors when sampling centers, combining them with random or farthest-point sampling.  \name\ permits us to use the locations of the $K$ highest-confidence bounding box predictions from the previous frame quite easily (Figure~\ref{fig:starnet_seed}): we can replace the last $K$ farthest-point-sampled (or random) center proposals for the current frame using the pose-corrected locations of the previous top $K$ detection bounding boxes from the prior frame.

We apply this method to Pedestrian detection (Table~\ref{table:previous detects}).  When using just 32 of the high-confidence predicted bounding box centers from the previous frame and a total of 384 centers, detection mAP on the validation set increases by over 10 absolute mAP, matching the (single frame) performance of sampling a total of 512 centers.  When we seed the 384 sampled centers with the top 192 detects from the previous frame, detection mAP improves by nearly 17 absolute mAP, or 40\% higher.  Surprisingly, when using 1024 total centers, using the best 512 previous detection centers improved mAP by about 3 mAP (2 mAP on the test set), showing that there is room for improvement even when sampling already covers much of the scene.

This experiment demonstrates that using \name\ enables research into smarter and efficient detection and tracking systems.  One could employ the use of a tracker to estimate the velocity of detected objects in order to more precisely predict where to ``look" in the next frame.

\begin{table}
\footnotesize
\begin{center}
\begin{tabular}{ c | c | c }
\hline
\# Previous Frame & \# Total Centers & Detection mAP \\
Detection Centers & & Pedestrians \\
\hline
0 & 384 & 41.8 \\
32 & 384 & 53.2 \\
192 & 384 & 58.0 \\
\hline
0 & 1024 & 66.8 \\
512 & 1024 & 69.7
\end{tabular}
\end{center}
\vspace{0.1cm}
\caption{\textbf{Previous frame detection centers are good centers to use in the current frame.}  \name\ enables using data-dependent centers from the detection outputs of the previous frame to improve detection performance in the current frame. Results reported on the \textit{Validation set}.}
\label{table:previous detects}
\vspace{-8pt}
\end{table}

\vspace{-5pt}
\section{Discussion}

In this work, we presented a non-convolutional detection system that operates on native point cloud data. The goal of the proposed method is to better match the sparsity of point cloud data, and also allow the system to be flexibly targeted across a range of computational priorities. We demonstrate that the resulting detector is competitive with state-of-the-art detection systems on the KITTI object detection benchmark \cite{geiger2013vision}, and can outperform a competitive convolutional baseline on the large-scale \datasetname.

The system allows for targeted computation, enabling the use of temporal context from detection outputs of prior frames. We show up to a 40\% relative improvement in mAP using prior frames to inform where to target computation for the current frame. We further demonstrate how in principle the detection system can target spatial locations without retraining nor sacrificing the prediction quality. For instance, depending on evaluation settings, a single trained pedestrian model can exceed the predictive performance of a baseline convolutional model by $\sim 48\%$ at a similar FLOPS; or, the same model may achieve the same predictive performance but with $\sim20\%$ of the FLOPS.

We foresee multiple avenues for further improving the fidelity of the system including: multi-sensor fusion with cameras
\cite{yang2018ipod, qi2018frustum,liang2019multi}, employing semantic information such as road maps to spatially target detections \cite{yang2018hdnet}, or restoring global context by removing conditional independence from each proposal \cite{xie2018attentional}. 
While we have focused this first work on relatively simple sampling methods for proposals, more expensive or learned methods may further improve the system \cite{faster_rcnn}. For example, one could learn a ranking function to order the relative importance of proposals for a self-driving planning system \cite{cohen1998learning,burges2005learning,cao2007learning}.
Finally, we are particularly interested in studying how this system may be amenable to object tracking
\cite{bertinetto2016fully, held2016learning,gordon2018re} as we suspect that because of the design, the computational demands may scale as the {\it difference} between successive time points as opposed to operating on the entirety of the scene \cite{feichtenhofer2017detect}.

\ifcvprfinal
\textbf{Acknowledgements:} We wish to thank Tsung-Yi Lin, Chen Wu, Junhua Mao, Henrik Kretzschmar, Drago Anguelov, George Dahl, Anelia Angelova and the larger Google Brain and Waymo teams for support and feedback.
\fi

{\small
\bibliographystyle{ieee_fullname}
\bibliography{paper}
}

\clearpage

\appendix

\begin{strip}
\begin{center}
{\Large {\bf Supplementary Material:\\ Targeted Computation for Object Detection in Point Clouds}}
\label{appendix_full_test_results}
\end{center}
\end{strip}

\section{Full Waymo Open Dataset Results}
\begin{wraptable}{l}{17cm}
\begin{center}
\begin{tabular}{|l|c|c|c|c|}
\hline
                                       & Overall   & 0-30m     & 30-50m    & 50m-Inf   \\ \hline
PointPillars Pedestrians mAP           & 60.0/54.0 & 68.9/66.4 & 57.6/52.9 & 46.0/37.0 \\ 
PointPillars Pedestrians mAPH          & 47.3/42.5 & 55.8/53.7 & 45.0/41.2 & 33.4/26.8 \\ 
\hline
StarNet Pedestrians mAP                & 70.1/63.2 & 78.6/75.7 & 67.9/62.7 & 57.2/46.1 \\ 
StarNet Pedestrians mAPH               & 35.6/32.1 & 40.3/38.9 & 34.5/31.8 & 28.0/22.6 \\ 
StarNet (directional) Pedestrians mAP  & 67.8/61.1 & 76.0/73.1 & 66.5/61.2 & 55.3/44.5 \\ 
StarNet (directional) Pedestrians mAPH & 59.9/54.0 & 67.8/65.2 & 59.2/54.5 & 47.0/37.8 \\ 
\hline
\hline
PointPillars Vehicles mAP              & 62.2/54.5 & 81.8/80.7 & 55.7/50.1 & 31.2/23.2 \\
PointPillars Vehicles mAPH             & 61.7/54.0 & 81.3/80.2 & 55.1/49.6 & 30.5/22.7 \\ \hline
StarNet Vehicles mAP                   & 64.7/56.3 & 83.3/82.4 & 58.8/53.2 & 34.3/25.7 \\
StarNet Vehicles mAPH                  & 45.5/39.6 & 62.0/61.3 & 35.9/32.5 & 20.5/15.4 \\
StarNet (directional) Vehicles mAP     & 61.5/54.9 & 82.2/81.3 & 56.6/49.5 & 32.2/23.0 \\
StarNet (directional) Vehicles mAPH    & 61.0/54.5 & 81.7/80.8 & 56.0/49.0 & 31.8/22.7 \\ \hline
\end{tabular}
\end{center}
\caption{Waymo Open Dataset \textit{Test set} results for StarNet versus a PointPillars \cite{lang2018pointpillars} baseline model. Every table item is the LEVEL\_1/LEVEL\_2 mean average precision (mAP) or heading weighted mean average precision (mAPH).}
\label{waymo_table_full}
\end{wraptable}

\end{document}